\documentclass[10pt,twocolumn,letterpaper]{article}

\usepackage[pagenumbers]{cvpr} 

\usepackage{graphicx}
\usepackage{amsmath}
\usepackage{amssymb}
\usepackage{booktabs}
\usepackage{algorithm}
\usepackage[ruled,vlined,algo2e]{algorithm2e}
\usepackage{algpseudocode}
\usepackage{bm}

\usepackage{multirow}
\usepackage[T1]{fontenc}
\usepackage{verbatim}
\usepackage{array}

\usepackage[pagebackref,breaklinks,colorlinks]{hyperref}

\usepackage[capitalize]{cleveref}
\crefname{section}{Sec.}{Secs.}
\Crefname{section}{Section}{Sections}
\Crefname{table}{Table}{Tables}
\crefname{table}{Tab.}{Tabs.}

\def\cvprPaperID{*****} 
\def\confName{CVPR}
\def\confYear{2023}

\begin{document}

\title{Normalizing Flow based Feature Synthesis for Outlier-Aware Object Detection}




\author{Nishant Kumar$^{1}$, Sini\v{s}a \v{S}egvi\'{c}$^{2}$, Abouzar Eslami$^{3}$, and Stefan Gumhold$^{1}$\\
$^{1}$TU Dresden, $^{2}$University of Zagreb - FER, $^{3}$Carl Zeiss Meditec AG}

\maketitle

\begin{abstract} 
Real-world deployment of reliable object detectors is crucial for applications such as autonomous driving. However, general-purpose object detectors like Faster R-CNN are prone to providing overconfident predictions for outlier objects. Recent outlier-aware object detection approaches estimate the density of instance-wide features with class-conditional Gaussians and train on synthesized outlier features from their low-likelihood regions. However, this strategy does not guarantee that the synthesized outlier features will have a low likelihood according to the other class-conditional Gaussians. We propose a novel outlier-aware object detection framework that distinguishes outliers from inlier objects by learning the joint data distribution of all inlier classes with an invertible normalizing flow. The appropriate sampling of the flow model ensures that the synthesized outliers have a lower likelihood than inliers of all object classes, thereby modeling a better decision boundary between inlier and outlier objects. Our approach significantly outperforms the state-of-the-art for outlier-aware object detection on both image and video datasets. 

\end{abstract}

\section{Introduction}
\label{sec:intro}
General purpose object detectors such as Faster R-CNN~\cite{ren2015faster} and Mask R-CNN~\cite{https://doi.org/10.48550/arxiv.1703.06870} deliver high performance for inlier images. However, in many real-world scenarios, such as autonomous driving~\cite{Choi_2019_ICCV}, plenty of unknown outliers (OD) naturally occur in an image or a video scene. Due to the co-existence of OD with the labeled inlier (ID) objects in the scene, object detectors confuse outliers with inliers. Therefore, reliable object-detection deployments require detecting such anomalies without degrading the performance of inlier object detection.

Many outlier detection approaches focus on multi-class image classification task by either performing outlier detection during inference~\cite{hendrycks2016baseline, liang2018enhancing, lee2018simple, liu2020energy, DBLP:conf/icml/SastryO20, huang2021importance} or training on real outlier data~\cite{hendrycks2018deep, mohseni2020self,DBLP:journals/corr/abs-2106-03917, bevandic22ivc}. 
However, such OD inputs are unaware of the decision boundary between inliers and outliers, resulting in an inaccurate model regularization.
A popular work~\cite{lee2018training} proposed a novel training scheme to generate synthetic outlier samples in the high-dimensional pixel space using Generative Adversarial Networks for outlier aware training of an image classification model. However, GANs are challenging to optimize and are likely to deliver insufficient coverage~\cite{lucas19neurips}. 
Moreover, the previous work~\cite{lee2018training} generates an entire image as an outlier sample, whereas in an object detection problem, both inliers and outliers can coexist in the same image space. 

\begin{figure}[t]
    \centering
    \vspace{-1em}
    \includegraphics[width=1.0\linewidth]{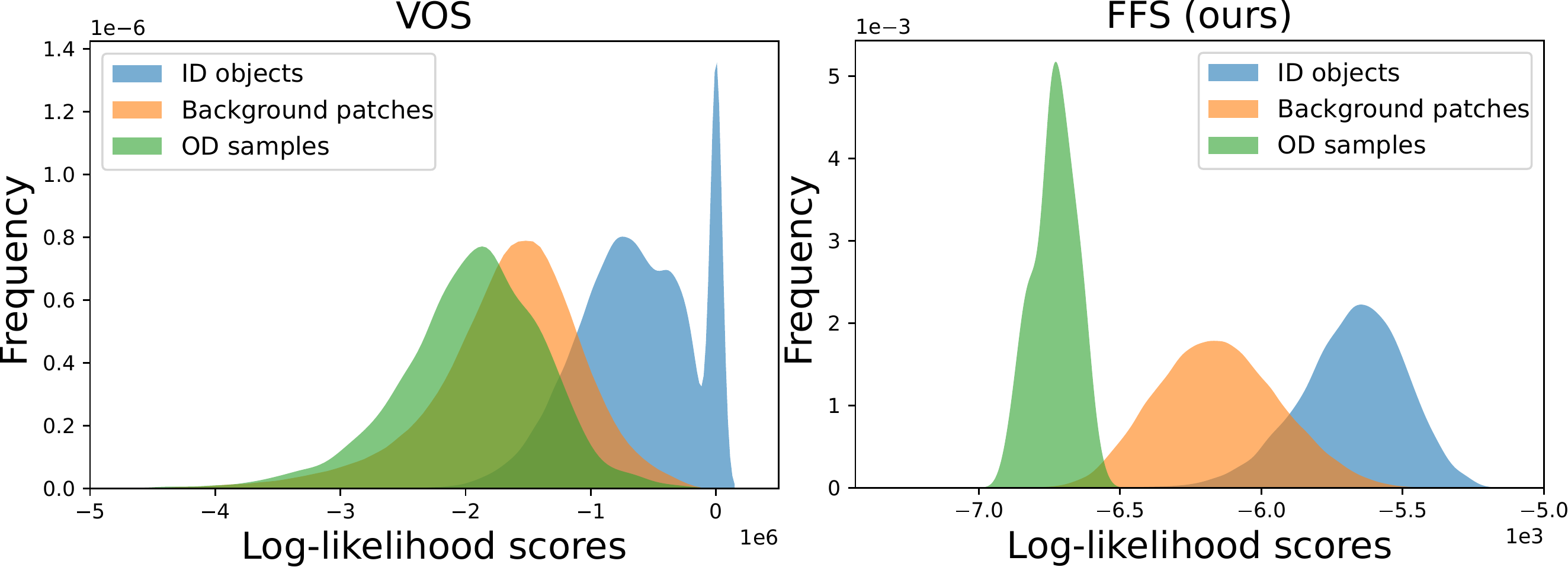}
    \caption{\small We compare distributions of log-likelihoods
for inliers (ID), background patches, and synthetic outlier (OD) samples by applying a pre-trained model on Pascal-VOC~\cite{DBLP:journals/ijcv/EveringhamGWWZ10} validation set. The left plot shows the likelihoods of class-conditional Gaussians from the official approach
of VOS~\cite{du2022vos}. The right plot shows the likelihoods recovered with our normalizing flow.
We observe that our flow
separates the three distributions
much better than VOS. }
    \label{fig:teaser}
    \vspace{-1.2em}
\end{figure}

Recently,~\cite{du2022vos} proposed to model the inlier features as class-conditional Gaussians and then synthesize OD features from the low likelihood region of the modeled distribution. Even though density is more easily estimated
in the feature space than in pixel space, the assumption of class-conditional Gaussian may not provide an accurate decision boundary between inliers and outliers.~Furthermore, synthesizing a low-likelihood OD sample from the Gaussian for class A does not guarantee that the sample is also of the low likelihood for the Gaussian of class B, as indicated in the left plot of Figure~\ref{fig:teaser}.
Recent work for object detection in video~\cite{du2022stud} extracts the background patches for uncertainty regularization by thresholding the dissimilarity with the reference inlier features.~However, the extracted background patches may lie far from the inlier-outlier decision boundary, leading to sub-optimal regularization of the model. 

We propose a novel approach for open-set object detection which we call Flow Feature Synthesis (\texttt{FFS}).~Our approach trains an invertible normalizing flow to map the data distribution of ID features of all object classes to a latent representation that conforms 
to a multivariate Gaussian distribution
with zero mean and diagonal unit covariance.~As a result, it ensures principled estimation of the complex distribution of inlier features from all classes. We synthesize outlier  features from low-likelihood regions of the learned distribution.~This enables outlier features to be near the ID-OD decision boundary, leading to more robust uncertainty regularization of the object detector.~In contrast, using generative adversarial models 
for the same task would be cumbersome
due to its inability to infer density of the generated samples,
Furthermore, variational autoencoders
can only infer the lower bound of the sample density~\cite{kingma2018glow}.


Specifically, \texttt{FFS} optimizes the normalizing flow model by maximum likelihood training on ID features. This training scheme enables the model to estimate the actual data distribution of the available ID features. Next, \texttt{FFS} utilizes the invertibility of the flow model to randomly sample from its latent space and generate synthetic features in the reverse direction of the model. 
The normalizing flow allows efficient and exact inference
of the distribution density 
in the generated samples.
Consequently, our approach can deliver
suitable synthetic outlier data
in fewer iterations than VOS~\cite{du2022vos}. 
It also requires fewer synthetically generated samples than VOS~\cite{du2022vos} to obtain OD features from the low-likelihood region of the modeled distribution by the flow.
Furthermore, we developed our end-to-end trainable $\texttt{FFS}$ framework to be effective on both image and video datasets. In contrast, previous works~\cite{du2022stud, du2022vos} proposed standalone strategies for each task. Our main contributions are: 

\vspace{-0.6em}
\begin{itemize}
\item We present a new outlier-aware object detection framework that utilizes Normalizing Flows to model the joint data distribution of inlier features. Invertibility of the flow allows 
efficient generation 
of synthetic outliers for effective uncertainty regularization. 
\vspace{-0.7em}

\item By mapping the data distribution of inlier features from all object classes to a multivariate Normal distribution in the flow's latent space, ~\texttt{FFS} ensures that an outlier sampled using the flow model is OD with reference to all ID classes. 
\vspace{-0.7em}

\item ~\texttt{FFS} achieves better OD detection performance while training faster than VOS~\cite{du2022vos} and STUD~\cite{du2022stud}, due to having to generate fewer synthetic samples. 
\vspace{-0.7em}

\item We show that our method achieves state-of-the-art performance in OD object detection while preserving the baseline ID detection performance for image dataset PASCAL-VOC~\cite{DBLP:journals/ijcv/EveringhamGWWZ10} and video datasets such as Youtube-VIS~\cite{DBLP:conf/iccv/YangFX19} and BDD100K~\cite{DBLP:conf/cvpr/YuCWXCLMD20}. 

\end{itemize}
\vspace{-1em}

\begin{figure*}[t]
    \centering
    \includegraphics[width=1.0\linewidth]{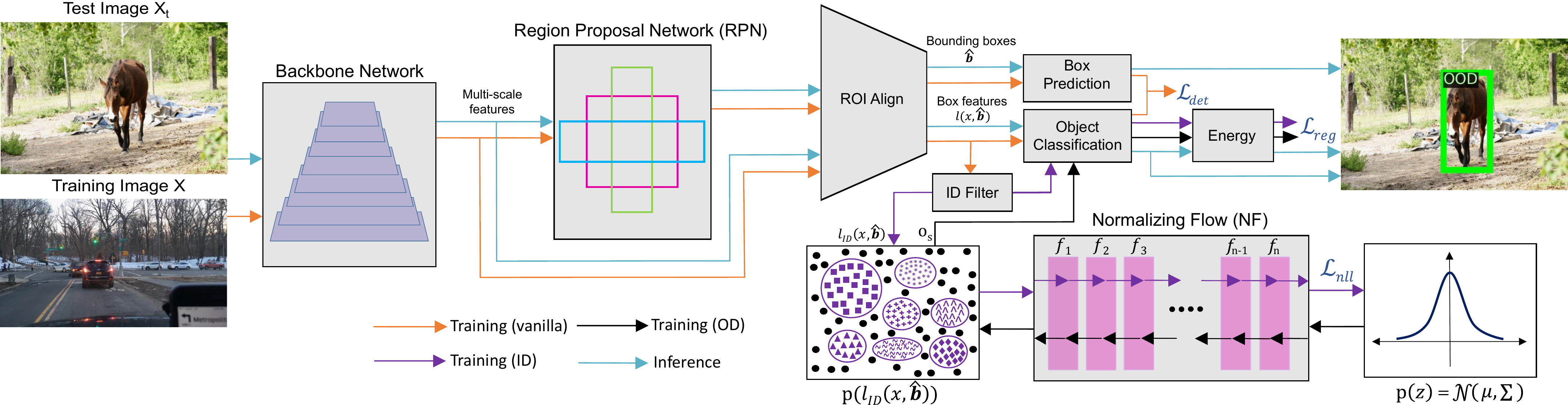}
    \vspace{-2em}
    \caption{\small \textbf{Overview of the} \texttt{FFS} \textbf{framework.} 
    The ID features $l_{ID}(x, \boldsymbol{\hat{b}})$ are selected for training the normalizing flow with the negative log-likelihood loss $\mathcal{L}_{nll}$. We recover outlier features $o_s$ by iterative sampling in the latent space of the normalizing flow. The classification logits of the ID features $l_{ID}(x, \boldsymbol{\hat{b}})$ and the outlier features $o_s$  are used to calculate the energy scores. The uncertainty loss $\mathcal{L}_{reg}$ encourages separation between the energy scores of inlier and outlier features. More details are given in Section~\ref{method}.}
    \label{fig:framework}
    \vspace{-1.2em}
\end{figure*}

\section{Related work}
The research in outlier-aware object detection is in the early stages compared to  works in vanilla object detection~\cite{girshickICCV15fastrcnn, https://doi.org/10.48550/arxiv.1703.06870, ren2015faster} or outlier-aware semantic segmentation~\cite{visapp21}. This section overviews prior approaches to outlier detection for image classification and proceeds to outlier-aware object detection for images and videos.
\vspace{-0.5cm}
\paragraph{Outlier-aware image classification.} Past approaches trained a classification model without knowledge about outliers and conducted OD detection during inference. For example,~\cite{hendrycks2016baseline} used simple softmax probability scores during inference to detect OD samples, while other works used Mahalanobis distance~\cite{lee2018simple}, rectified activations~\cite{sun2021react}, KL divergence~\cite{huang2021importance} and Gram Matrices~\cite{DBLP:conf/icml/SastryO20} instead of softmax scores to detect such samples. On the other hand, ODIN~\cite{liang2018enhancing} and Generalized ODIN~\cite{hsu2020generalized} performed perturbations to the test examples to enhance the performance of softmax function for OD detection. Recently, energy function~\cite{liu2020energy, lin2021mood, wang2021canmulti} has performed better for distinguishing ID and OD samples than softmax scores.

Several other methods~\cite{hendrycks2018deep,mohseni2020self,hendrycks2020augmix,DBLP:conf/nips/HendrycksMKS19,tack2020csi} proposed to regularize image classifiers by exposing them with outlier data during  self-supervised training while a GAN based approach~\cite{lee2018training} synthesizes outliers in pixel space and uses it for outlier-aware image classification. However, such synthetic outliers are sampled from imprecise decision boundaries. Additionally, these approaches may not be applied to an object detection task since outliers defined by such methods are in the entire pixel space of the input image. In contrast, an image may contain both inlier and outlier objects simultaneously, and a trustworthy object detection model should detect such scenarios.  

\vspace{-0.5cm}
\paragraph{Outlier-aware object detection for images.}
One of the early open-set object detection works~\cite {DBLP:conf/wacv/DhamijaGVB20} highlighted the importance of adding an extra background class for unknown objects, though object detection inherently rejects unknown object proposals. Several works improve the background detection in open-set conditions~\cite{DBLP:conf/icra/MillerDMS19, DBLP:conf/icra/MillerNDS18,DBLP:conf/wacv/0003DSZMCCAS20,DBLP:journals/corr/abs-2108-03614} by using Bayesian techniques, such as Monte Carlo-Dropout~\cite{gal2016dropout}. However, such methods have high latency during inference which could be sub-optimal for real-world applications. Others~\cite{DBLP:journals/corr/abs-2101-05036,DBLP:journals/corr/abs-2107-04517} presented uncertainty regularization for localization regression but did not examine outlier-aware object detection with a focus on classification head.
Works like~\cite{DBLP:journals/corr/abs-2103-02603} used the background patches as OD samples for model regularization, but the performance might be low as the samples may be far from the decision boundary. VOS~\cite{du2022vos} applies energy-based regularization~\cite{liu2020energy} to synthetic OD features generated 
by class-conditional Gaussians. Instead, our method map features from all ID classes to a Gaussian prior with a normalizing flow model, and thus achieves a tighter decision boundary and improved performance.
\vspace{-0.5cm}
\paragraph{Outlier-aware object detection for videos.}
Several works aim to identify anomalous events on the object and frame level. An object-level approach~\cite{DBLP:conf/cvpr/DoshiY20b} uses the $k$-nearest neighbors algorithm to detect anomalous objects while~\cite{DBLP:conf/cvpr/IonescuKG019} use $k$-means to cluster inlier features and train $k$ binary classifiers so that each cluster can view other clusters as an outlier. A test object is labeled an outlier if the highest classification score is negative. Frame-level approaches~\cite{DBLP:conf/cvpr/LiuLLG18, DBLP:conf/wacv/RavanbakhshNMSS18} predicted future frames that
preserve the events of the previous frames by ensuring the optical flow of the predicted frame remains consistent with ground-truth. In contrast, events with a significant difference between prediction and ground-truth are labeled outliers.
However, no previous work synthesizes outliers from video frames for model regularization. Recent work~\cite{DBLP:journals/corr/abs-2104-08381} introduced self-supervised learning to find the most distinct object in the next video frame given an inlier object in the previous frame, leading to more knowledge about unseen data domains. STUD~\cite{du2022stud} uses the inliers in the first frame to select the outliers in the following frames for model regularization. However, this may deliver outlier patches that are very far away from the decision boundary and thus lead to inferior model regularization. To summarize, no single outlier-aware object detection framework exists for images and video with an effective regularization that models a precise inlier manifold. 

\section{Method}
\label{method}

Outlier-aware object detection problem is much more complex than detecting outliers for an image classification task since a real-world image  may consist of both inlier and outlier objects. Let us define an input \({x} \in \mathcal{X}\) with $\mathcal{X}:= (x_1, x_2,..., x_N)$ being the training dataset consisting of $N$ images or video frames, ground-truth class labels for each ID object as \({y} \in \mathcal{Y} \) with $\mathcal{Y}:= \{1, 2,..., K\}$ for $K$ classes and the coordinates of ground-truth bounding boxes \(\boldsymbol{b} \in \mathcal{B} \). 
Then, an end-to-end object detector with parameters $\theta$ treated as a conditional probability distribution  estimates bounding boxes as  \(p_{\theta}({\boldsymbol{b}|x})\) and the object class as \(p_{\theta}({y|x,\boldsymbol{b}})\).  Figure~\ref{fig:framework} shows the training and inference scheme of the $\texttt{FFS}$ framework. The backbone network extracts the feature maps at different resolutions, and the Region Proposal Network (RPN) converts these multi-resolution feature maps into object proposals. The ROI align module classifies object proposals and provides a vector of bounding boxes, $\boldsymbol{\hat{b}}$, and fixed-size box features, $l(x, \boldsymbol{\hat{b}})$, of both ID (inlier) and background patches. Note that $l(x,\boldsymbol{\hat{b}})$ is in a lower dimensional space than $x$. The box features $l(x, \boldsymbol{\hat{b}})$ and bounding boxes $\boldsymbol{\hat{b}}$ are used to train the object classification and box prediction heads according to the object detection loss $\mathcal{L}_{det}$, given the ground-truth ID classes and box coordinates, respectively. The box features from $l(x,\boldsymbol{\hat{b}})$ with the ground-truth class as background are filtered out to keep only ID features $l_{ID}(x,\boldsymbol{\hat{b}})$. 

\subsection{Overview}
\label{overview}
Our framework \texttt{FFS} is split into three stages: the training procedure, the threshold estimation for OD detection, and the model inference as summarized in Algorithm~\ref{alg:algo}. During the training stage, we compute the standard object detection loss $\mathcal{L}_{det}$ via the chosen general-purpose object detector. We define maximum-likelihood training objective $\mathcal{L}_{nll}$  to train our normalizing flow module on inlier features $l_{ID}(x, \boldsymbol{\hat{b}})$. Subsequently, we sample synthetic outliers $o_s$ from the flow model. The details related to the normalizing flow and the procedure to synthesize outlier features are given in Section~\ref{method:Normalizing Flow}. We then compute energy scores for inliers and outliers for the model regularization using $\mathcal{L}_{reg}$. Section~\ref{method:regularization} describes the details about model regularization, and Section~\ref{method:inference} discusses the overall training objective of the $\texttt{FFS}$ framework. We use the trained $\texttt{FFS}$ model for threshold estimation to perform OD detection. We fix the energy-based threshold $\xi$ such that the model correctly detects 95\% of inlier objects in the validation set. During inference, given an object $x_t$ in a test image, the trained $\texttt{FFS}$ model provides a bounding box for the object as $b_t$. The decision of whether the object is an inlier or an outlier lies on the energy $E(h(l(x_t,b_t);\theta))$.
Given an energy threshold $\xi$, we assign $x_t$ as an inlier if $E(h(l(x_t,b_t);\theta)) < \xi$ and OD if $E(h(l(x_t,b_t);\theta))  \geq \xi$.  We obtain object class labels for predicted inliers with the softmax confidence score and bounding box. We  label the bounding box for objects detected as outliers as OOD.

\vspace{-0.7em}
\begin{algorithm}[H]

\SetAlgoLined
\textbf{Input:} Training data $\left\{\left(\*x_{i}, \mathbf{b}_i,{y}_{i}\right)\right\}_{i=1}^{N}$, vanilla object detector, normalizing flow $f$ and binary classifier $\Phi$ with  parameters $\theta$, $\gamma$ and $\psi$ respectively, the loss weights $\alpha$ and $\beta$.

\textbf{Output:} Object detector with trained parameter $\theta$.\\
\While{train}{

1.~Input training data to \texttt{FFS} model and compute $\mathcal{L}_{{det}}$.  

2.~Input the ID features $l_{ID}(x,\boldsymbol{\hat{b}})$ to the flow model $f$ and obtain  $\mathcal{L}_{{nll}}$ using Eq~\ref{eq2}. 

\eIf {$\texttt{i} \leq \ \textit{iterations}$}{

3.~Update $i$ and the parameters $\theta$ and $\gamma$ by backpropagation of the total loss in Eq~\ref{eq:all_loss} with $\alpha = 0$.
}
 {
4.~Generate synthetic features $g_k$ by sampling $z_k$ from flow's latent space with $g_k = f^{-1}(z_k)$.

5.~Select $o_s$ from $p_\gamma(g_k)$'s low-likelihood region.

6.~Calculate energy scores $E(h(l_{ID}(x,\boldsymbol{\hat{b}});\theta))$ and $E(h(o_s;\gamma))$ using the classification head $h$.

7.~Compute regularization loss $\mathcal{L}_{reg}$ using Eq~\ref{eq4}. 

8.~Update $i$ and the parameters $\theta$, $\gamma$ and $\psi$ by backpropagation of the total loss in Eq~\ref{eq:all_loss}.}
}
Fix threshold $\xi$ when 95\% of inliers are correctly detected.\\
\While{eval}{
  1.~Calculate the energy score of a test object.
  
  2.~Label it as an outlier or an inlier based on $\xi$.}
 \caption{\texttt{FFS}\xspace: Normalizing Flow based Feature Synthesis for Outlier-Aware Object Detection}
 
 \label{alg:algo}
\end{algorithm}

\vspace{-0.7em}
\subsection{Normalizing Flow for feature synthesis}
\label{method:Normalizing Flow}
For simplicity, we will denote the ID features $l_{ID}(x,\boldsymbol{\hat{b}})$ as $l \in \mathcal{L}$. In our framework, the normalizing flow $f$ with parameters $\gamma$ is a sequence of $M$ invertible bijective mappings implemented as affine coupling layers.   The flow \(f:\mathcal{L} \rightarrow \mathcal{Z}\) transforms the complex data distribution of the features $l$ to a multivariate Gaussian in its latent space \({z} \in \mathcal{Z}\)  with \(p({z}) = \mathcal{N}_d( \boldsymbol \mu, \boldsymbol \Sigma)\) where $d$ is the number of feature vector components, and $\boldsymbol \mu$ is zero mean with $\boldsymbol \Sigma$ as the unit variance. The bijective mapping ensures that the input and the latent space share the same dimensionality $d$ such that \(f:\mathbb{R}^d \rightarrow \mathbb{R}^d\). Each coupling layer transforms the feature vector components by scaling $s$ and translation $t$, which are learnable neural networks~\cite{kingma2018glow}. Such a transformation is expressive and, at the same time, easily invertible with high efficiency in computing Jacobian determinants. 

\vspace{-1.2em}
\subsubsection{Maximizing the likelihood of ID features} 
 Given the above formulation of our normalizing flow $f$, which transforms input ID features $l$ into $z$ such that $z$ = $f(l)$, we aim to maximize the log-likelihood of recovering features $l$ with respect to $f$.  
To achieve this, we have to compute the Jacobian matrix of $f(l)$ with respect to the feature vector components~\cite{kingma2018glow}. Let $f_i(l)$ be the components of $f$ in latent space and $l_j$ the components of feature space. The entries of the Jacobian matrix are defined as
$J_{ij}^{f,l} =  \frac{\partial f_i(l)}{\partial l_j}$ where $i,j \in {1,...,d}$.
According to the change of variables formula, the posterior likelihood \(p_{\gamma}(l)\) can be described as:

 

\begin{equation}
p_{\gamma}(l) = p(f(l)) *
  \left|\det J^{f,l}\right|
 \label{eq1}  \end{equation}


To obtain the log-likelihood of the posteriors, we take a logarithm on both sides of the Eq~\ref{eq1}. As the training of a  neural network requires minimization of a loss function, we construct $f$ by finding $\gamma$ that  maximizes $\log(p_{\gamma}(l))$ by minimizing the negative log-likelihood $\mathcal{L}_{nll}$ objective:

\vspace{-1.5em}
\begin{equation}
\mathcal{L}_{nll}(l;\gamma) =  \frac{1}{N}  \sum_{i=1}^{N}  -\log(p_{\gamma}(l(x_i,b(x_i))))
\label{eq2} 
\end{equation}

Normalizing flows are known 
for stable and exact training according to the
maximum likelihood objective in Eq~\ref{eq2}.

\vspace{-1.0em}
\subsubsection{Random sampling from the latent space} 
We train the flow model $f$ so that its parameters $\gamma$ learn the distribution of inlier features for a fixed number of training iterations. Next, we randomly sample $k$ samples $z_k$ from the latent distribution $p(z)$ and propagate it in the reverse direction of $f$ to generate synthetic samples $g_k$ where $g_k = f^{-1}(z_k)$ during active training.
As the flow model learned the data distribution $p_\gamma(l)$ via the maximum likelihood training using Eq~\ref{eq2}, the samples $g_k$ are synthesized from  $p_\gamma(l)$.  
Such generation procedure discourages mode collapse, common in generative adversarial models. Hence, we retrieve the exact data distribution of the features $l$ in synthesized samples $g_k$. After that, we compute the log-likelihood scores of each of the $k$ samples $g_k$ using Eq~\ref{eq1}. Reasonably, the synthesized $g_k$ contains samples with a high likelihood of being obtained from $p_\gamma(l)$ and should lie well inside the boundary of the inlier distribution $p_\gamma(l)$. Nevertheless, there should also be samples in $g_k$ with lower likelihood scores that lie near or  away from the boundary of the distribution $p_\gamma(l)$. 


\vspace{-1.0em}
\subsubsection{Rejection Sampling based Outlier Synthesis} 
\begin{figure}[!htb]
    \centering
    \begin{subfigure}[ht!]{0.235\textwidth}
    \includegraphics[width=1.0\linewidth]{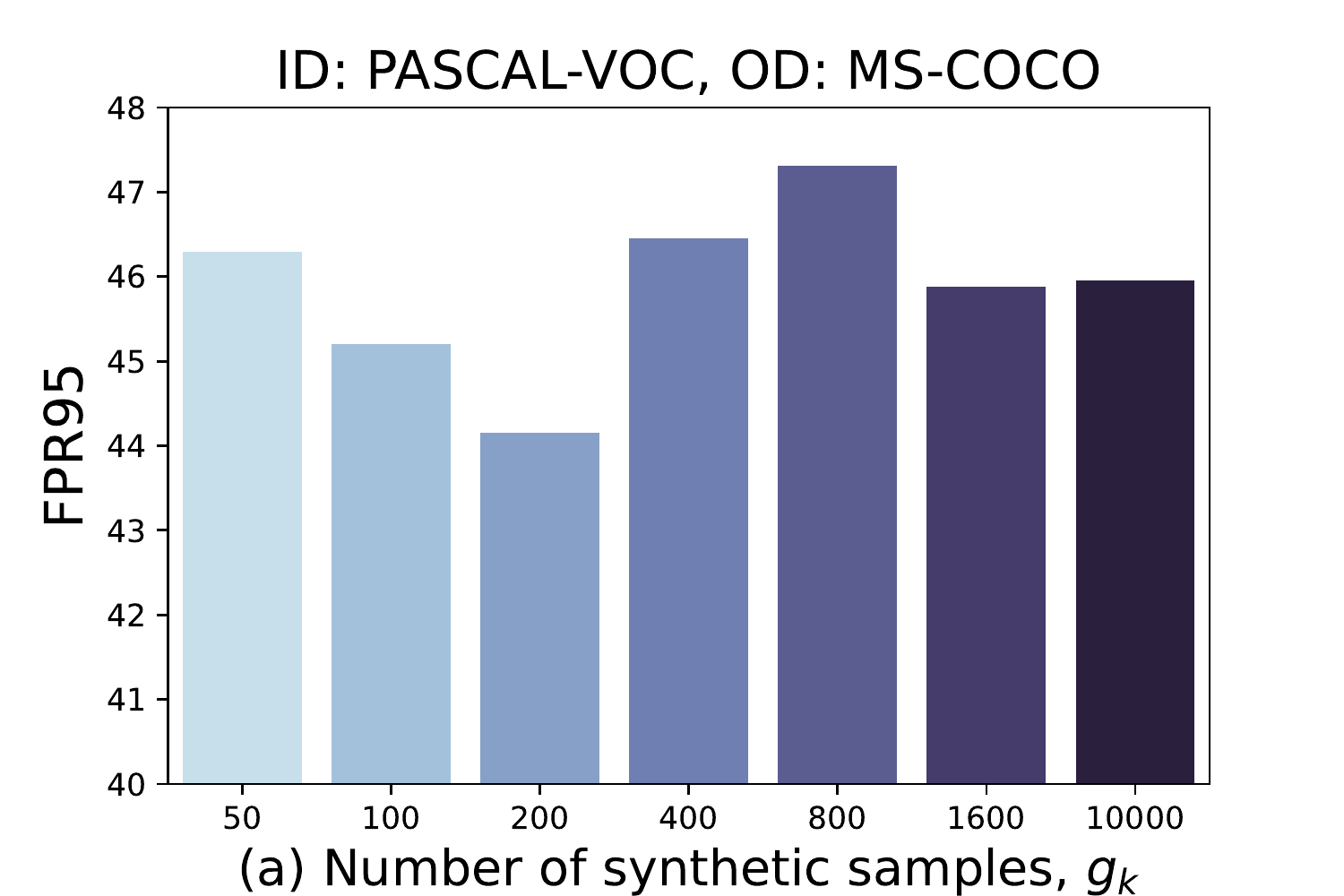}
    \end{subfigure}
    \begin{subfigure}[ht!]{0.235\textwidth}
    \includegraphics[width=1.0\linewidth]{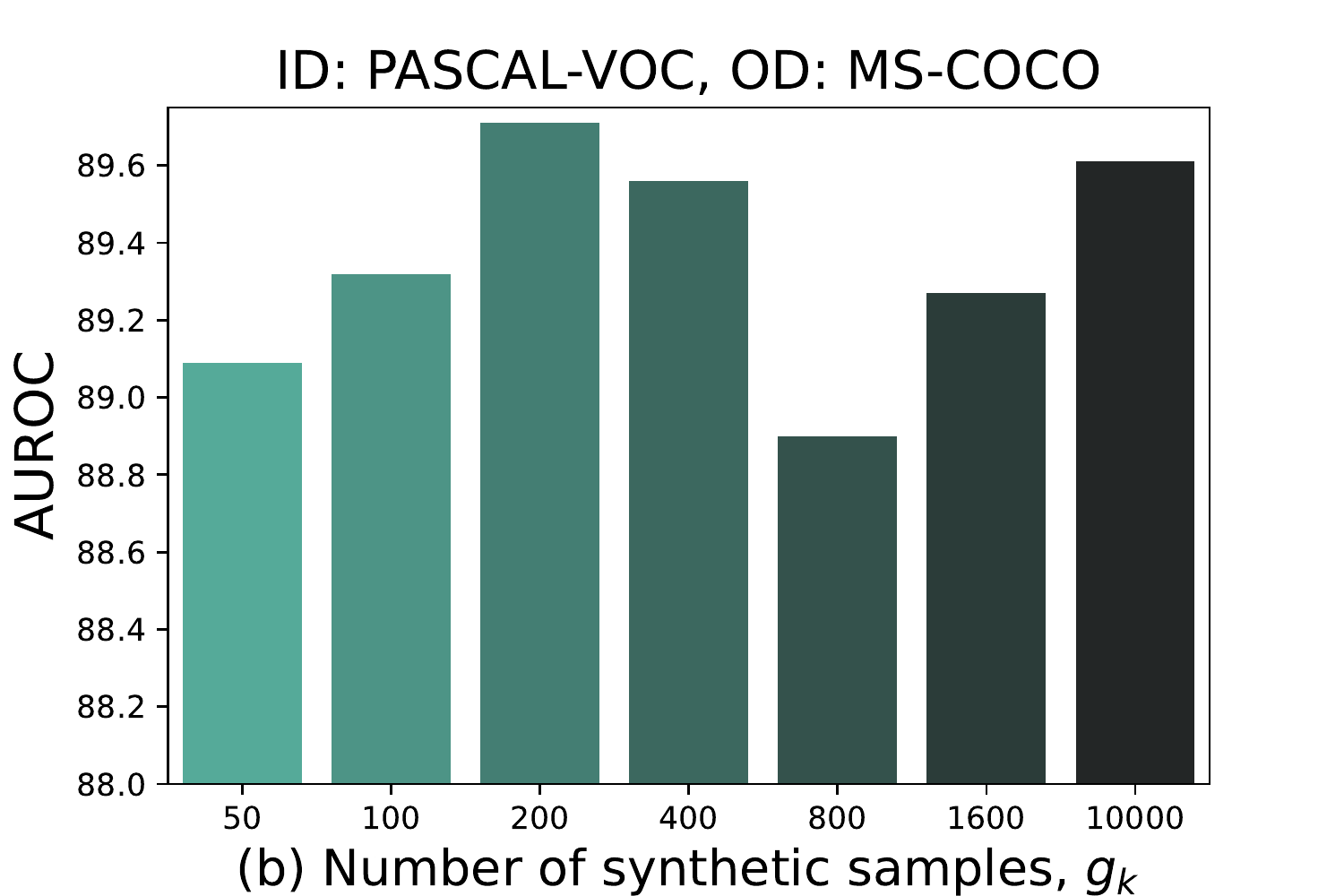}
    \end{subfigure}
    \begin{subfigure}[ht!]{0.235\textwidth}
    \includegraphics[width=1.0\linewidth]{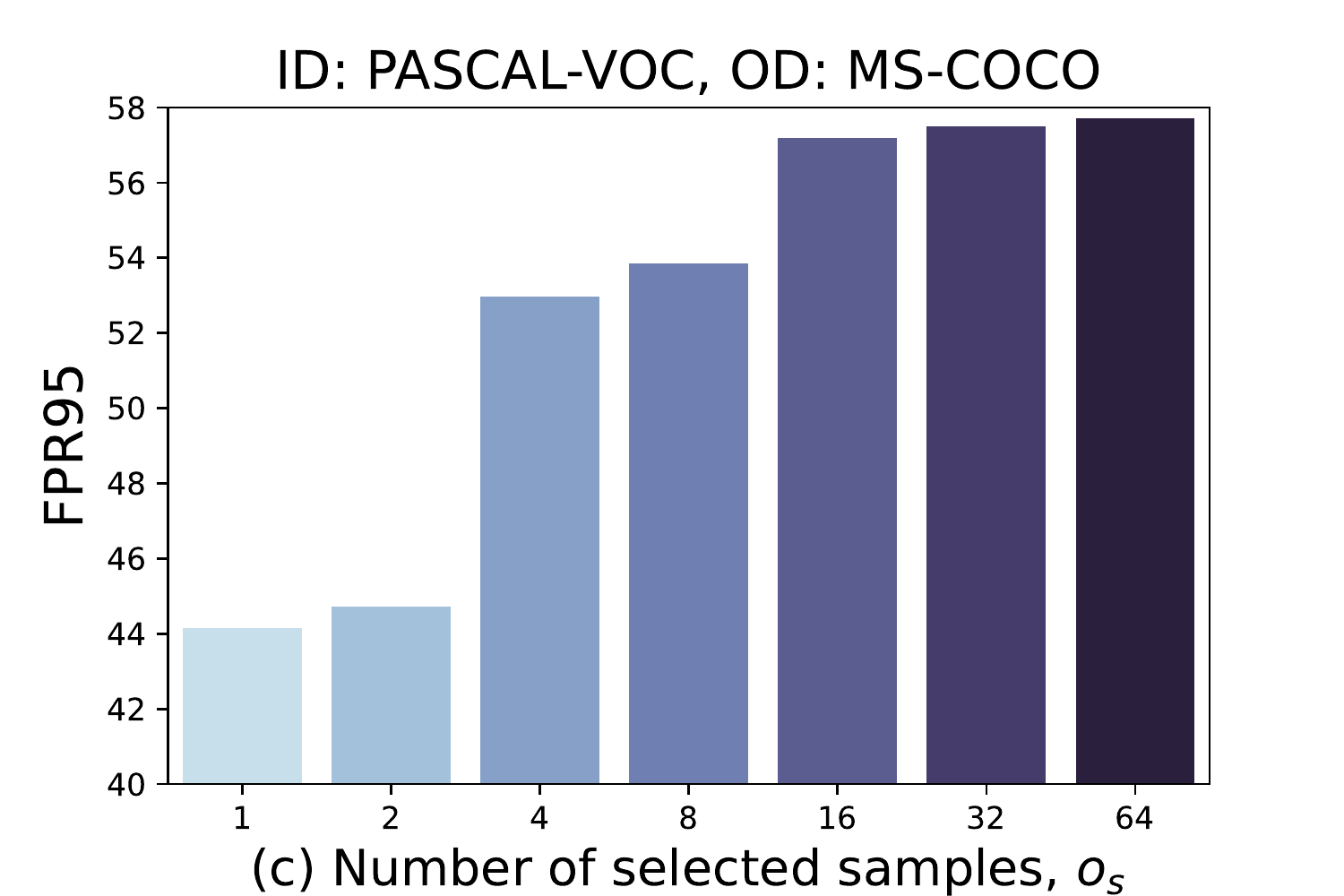}
    \end{subfigure}%
        \begin{subfigure}[ht!]{0.235\textwidth}
    \includegraphics[width=1.0\linewidth]{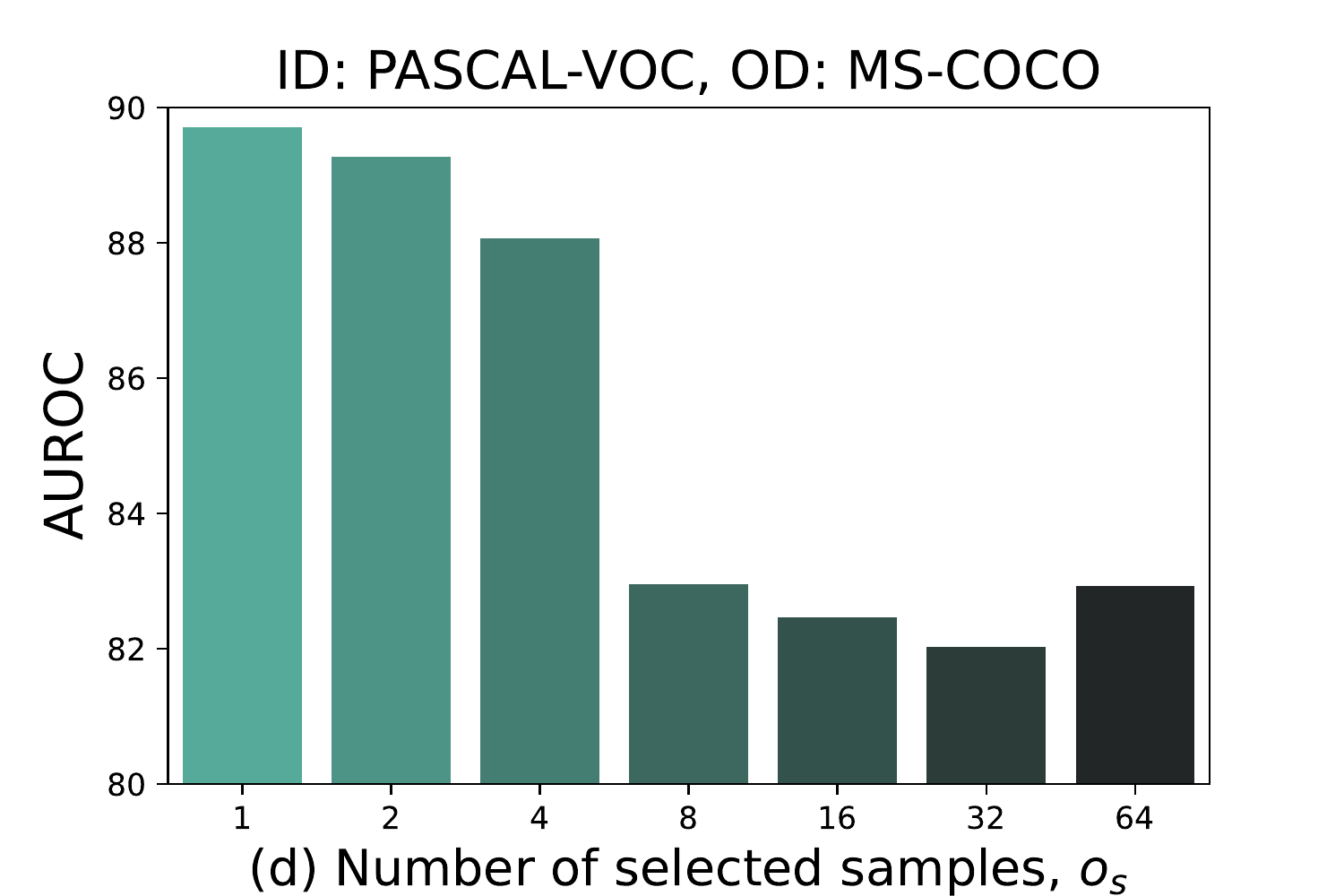}
    \end{subfigure}
    \caption{\small Validation of synthetic samples $g_k$ and selected outliers $o_s$ on OD detection. $s = 1$ in (a)-(b) and $k = 200$ in (c)-(d).}
    \label{fig:sampling}
    \vspace{-0.4em}
\end{figure}

 It is pivotal to sample useful synthetic outliers for an effective model regularization for OD detection. To achieve this, our first approach involves rejection sampling. We select $s$ samples as outliers $o_s$ from the low-likelihood region of the data distribution  $p_\gamma(g_k)$, where $o_s \subset g_k$ and $s$ being much smaller than $k$. The motivation of such an approach is to obtain synthetic outliers from near the decision boundary and outwards in the exterior outlier space. Figure~\ref{fig:sampling}~(a)-(b) shows that generating more synthetic samples $g_k$ and selecting a single outlier $o_s$ with the least likelihood (where $s =1$) improves the OD detection. This is because generating more samples $g_k$ leads to the selection of a low-likelihood synthetic outlier $o_s$ that genuinely lies in the outlier space, resulting in better model regularization.
 
  \vspace{-0.9em}
 Eventually, however, the performance peaks as generating an even higher number of synthetic samples $g_k$ may lead to the selection of an outlier $o_s$ that is too far away in the outlier space, resulting in a model that is unaware of the decision boundary. Hence, by varying $g_k$, our rejection sampling approach can obtain outliers from the decision boundary and from the outlier region away from the boundary.  Figure~\ref{fig:sampling} (c)-(d) show the results when the number of selected outliers $o_s$ is varied while keeping the number of the generated synthetic samples $g_k$ fixed. The selection of more outliers $o_s$ leads to a degradation in the OD detection performance as some selected outliers with high log-likelihood scores lie well within the inlier manifold leading to sub-optimal model regularization. 

\vspace{-1.0em}
\subsubsection{Projection Sampling using Langevin Dynamics} 
We investigate whether our model can generalize the outlier space by using synthetic outliers sampled only close to the decision boundary while still being effective with OD detection. To accomplish this, we perform projection sampling based on Stochastic Gradient Langevin Dynamics (SGLD) to sample outliers from our flow model. Under the approach, we generate $s$ samples as outliers $o_s$ directly without needing rejection and require a likelihood threshold to define the boundary of the inlier feature distribution. We fix this threshold as the log-likelihood score $\delta$ of the inlier, which is least likely to be obtained from the inlier data distribution $p_\gamma(l)$. Next, we propagate the average log-likelihood score of the synthetic outliers $\log p_{\gamma}(o_s)$ in the inverse direction of the flow without updating its parameters and obtain the gradients $\frac{\partial\log p_{\gamma}(o_s)}{\partial o_s}$. Subsequently, the outliers $o_s$ are updated as:
\vspace{-0.2em}
\begin{equation}
o_s := o_s - \tau \frac{\partial\log p_{\gamma}(o_s)}{\partial o_s}
\label{eq3}
\end{equation}
where $\tau$ is the step size of the gradient descent. We end this iterative process when the average log-likelihood of the updated $o_s$ is equal to or lower than $\delta$. Hence, the updated outliers $o_s$ are precisely projected to the near decision boundary of the inlier manifold where the least-likelihood inlier is located. The results of our projection sampling approach are shown in Figure~\ref{fig:lineplots} (f) and discussed in Section~\ref{langevin}.


\subsection{Energy-based model regularization}
\label{method:regularization}
We obtained the outliers $o_s$ by sampling from the data distribution $p_\gamma(l)$, requiring no actual outliers during training. The classification head $h$ now maps inlier features $l$ and outliers $o_s$ to $K+1$ classification logits as $h(l;\theta)$ and $h(o_s;\gamma)$, respectively. Note that $h(l;\theta)$ is influenced by all $\theta$ parameters, whereas $h(o_s;\gamma)$ depends on the flow parameters $\gamma$ and the parameters of the classification head. Recently,~\cite{liu2020energy,du2022vos} used energy function rather than softmax probabilities to differentiate inlier and outlier samples better. 
We follow~\cite{du2022vos}  and apply the energy function on the classification logits $h(l;\theta)$ and $h(o_s;\gamma)$ using $E(h(.))=- T\cdot \log\sum^{K} w_K \cdot e^{h_K(.)/T}$ where $T$ is a temperature parameter of the energy function and weight $w$ is learned to overcome class imbalances. We then pass the energy scores through a binary classifier $\Phi$ with parameters $\psi$ and  define the softmax output as $p_l$  when $E(h(l;\theta))$ is the input  and $p_o$ when $E(h(o_s;\gamma))$ is the input.
 Finally, we  use binary cross-entropy (BCE) based regularization loss $\mathcal{L}_{reg}$:
\vspace{-0.6em}
\begin{equation}
\begin{aligned}
     \mathcal{L}_{{reg}}&=  \frac{1}{N} \sum_{i=1}^{N} -(\log(p_l) + \log(1 - p_o))
     \label{eq4}
\end{aligned}
\end{equation}
 The Eq~\ref{eq4} shapes contrasting inlier and outlier energy surfaces by assigning low energy for inliers and high energy for outliers. 
%


\subsection{Overall Training Objective}
\label{method:inference}
We describe the standard object detection loss $\mathcal{L}_{det}$ as a combination of losses for object classification given the ground-truth inlier class labels $y$ and bounding box regression given the ground-truth box coordinates $\boldsymbol{b}$. Based on the formulation of the negative log-likelihood loss $\mathcal{L}_{nll}$ and the regularization loss $\mathcal{L}_{reg}$  from Eq~\ref{eq2} and \ref{eq4} respectively, we merge all these loss terms in our overall training objective: 
\begin{equation}
    \min _{\theta, \gamma, \psi} \mathbb{E}_{(\*x, \mathbf{b}, {y}) \sim \mathcal{X}, \mathcal{B}, \mathcal{Y}}~~\left[\mathcal{L}_{det}+ \beta \cdot \mathcal{L}_{nll} +\alpha \cdot \mathcal{L}_{reg}\right],
    \label{eq:all_loss}
\end{equation}

where $\alpha$ and $\beta$ are the weights of $\mathcal{L}_{reg}$ and  $\mathcal{L}_{nll}$ respectively. The $\mathcal{L}_{reg}$  loss acts as an adversary to $\mathcal{L}_{nll}$ loss as $\mathcal{L}_{nll}$ tries to maximize the likelihood of obtaining inlier features while $\mathcal{L}_{reg}$ forces the model to synthesize outliers away from inliers. Thus, such a training scheme encourages the generation of synthetic samples at the decision boundary resulting in outlier awareness of the object detector.


\section{Experiments}
This section evaluates our $\texttt{FFS}$ framework for its effectiveness with outlier-aware object detection. Section~\ref{impl} provides the details related to the image and video datasets, the neural network architecture, and the performance metrics. Section~\ref{sota} shows the results obtained with $\texttt{FFS}$ and compares it with other related approaches. Finally, Section~\ref{ablation} contains ablation studies to evaluate the performance of our approach in varied experimental settings. 

\subsection{Implementation details}
\label{impl}

\textbf{Datasets.}\label{data} We train our $\texttt{FFS}$ model on the publicly available image dataset PASCAL-VOC 2012~\cite{DBLP:journals/ijcv/EveringhamGWWZ10} containing 20 object categories as ID and use two OD datasets, namely MS-COCO~\cite{lin2014microsoft} and OpenImages~\cite{kuznetsova2020open} to evaluate the performance. Additionally, we train \texttt{FFS} on two video datasets, namely Berkeley DeepDrive (BDD100K)~\cite{DBLP:conf/cvpr/YuCWXCLMD20} and Youtube-Video Instance Segmentation (Youtube-VIS) 2021~\cite{DBLP:conf/iccv/YangFX19}. For the $\texttt{FFS}$ trained on video datasets, we infer on two OD datasets, namely MS-COCO~\cite{lin2014microsoft} and nuImages~\cite{DBLP:conf/cvpr/CaesarBLVLXKPBB20}. We use the pre-processed datasets by VOS~\cite{du2022vos} and STUD~\cite{du2022stud} so that the object categories in the OD dataset are independent of the object categories in the ID dataset.

\textbf{Network architecture.}
\label{model} We adopt the Faster R-CNN model~\cite{ren2015faster} for object detection provided in the publicly available Detectron2 framework~\cite{wu2019detectron2}. We use RegNetX-4.0GF~\cite{DBLP:conf/cvpr/RadosavovicKGHD20} as the backbone architecture due to its superior inlier detection performance. We employ Glow~\cite{kingma2018glow} as our normalizing flow model after observing lower performance for other flow architectures (details in Section~\ref{ablation}).

\textbf{Metaparameters.}
\label{hyp}
We run our experiments using Python 3.8.6 and PyTorch 1.9.0 with each ID dataset trained on four NVIDIA A100-SXM4 GPUs. We start the training of the $\texttt{FFS}$ framework by turning off uncertainty regularization loss $\mathcal{L}_{reg}$ for a fixed number of iterations. This enables the flow model to learn the data distribution of inlier objects from all object classes. Across datasets, we found the loss weightages $\beta$ = $\mathrm{10}^{-4}$ and $\alpha$ = $0.1$ as reasonable values for stable end-to-end training of the $\texttt{FFS}$ framework. In Section~\ref{ablation}, we perform an exhaustive meta-parameter study.

\textbf{Evaluation metrics.}
\label{metrics}
We compute the energy scores from $K$ classification logits of the inlier and outlier objects in the validation set. We evaluate these scores
according to the standard outlier detection metrics.
These metrics are Area Under the Curve Receiver Operating Characteristic (AUROC $\uparrow$) and False Positive Rate at 95$\%$ True Positive Rate (FPR95 $\downarrow$),  where True Positive is the correct detection of an inlier. For ID detection, we first compute Intersection over Union (IoU) between predicted bounding boxes and ground-truth boxes. Then, we compute the average precision (AP) for the standard range of IoU thresholds. Finally, the mean of the average precision over all known classes (mAP $\uparrow$) measures ID detection performance.

\subsection{Comparison with the State-of-the-Art}
\label{sota}
 Tables~\ref{tab:sota_image}~and~\ref{tab:sota_video} compare our method with other approaches on image and video datasets, respectively. 
  We use the results provided by VOS~\cite{du2022vos} and STUD~\cite{du2022stud} to show the performance of the compared approaches.
 Some evaluated approaches (except VOS and STUD) were developed for image-wide OD detection. However, these approaches can also be used to compare an outlier-aware object detection model since classification is one of the object detection tasks. No real outlier datasets were used during training to compare these approaches with our method. 
 
 \begin{table}[!htb]
\centering
\scalebox{0.90}{\begin{tabular}{l|c|c|c|c}
    \toprule
    \textbf{ID}  & 
   {\textbf{Method }}  &\textbf{FPR95} $\downarrow$  & \textbf{AUROC} $\uparrow$ & \textbf{ mAP} $\uparrow$ \\ \hline
    && \multicolumn{2}{c|}{OD: MS-COCO / OpenImages}  \\ 
    \cline{3-4} 
    \multirow{5}{*}{ \rotatebox[origin=l]{90}{\footnotesize PASCAL-VOC}} 
    & GAN~\cite{lee2018training}&	60.93 / 59.97 & 83.67 / 82.67 & 48.5 \\
      & Energy~\cite{liu2020energy}
    & 56.89 / 58.69 & 83.69  / 82.98& 48.7\\
  & CSI~\cite{tack2020csi}& 59.91 / 57.41 &81.83 / 82.95 & 48.1 \\
        &  VOS~\cite{du2022vos}&	47.77 / 48.33 & 89.00 / 87.59 & 51.5 \\
        & FFS (ours)  & \textbf{44.15} / \textbf{45.08} & \textbf{89.71} / \textbf{88.29}  & \textbf{51.8}  \\
        \bottomrule
\end{tabular}}
        \vspace{-0.2cm}
        \caption[]{\small Main results on image datasets
with Faster R-CNN model and RegNetX-4.0GF~\cite{DBLP:conf/cvpr/RadosavovicKGHD20} as the backbone for VOS and \texttt{FFS}.
 }
        \label{tab:sota_image}
        \vspace{-0.4em}
\end{table}

\vspace{-0.5em}
\begin{table}[!htb]
\centering
\scalebox{0.92}{\begin{tabular}{l|c|c|c|c|c}
    \toprule
    {\textbf{ID}}  & {\textbf{OD}}  &
   {\textbf{Method}}  &\textbf{FPR95} $\downarrow$  & \textbf{AUROC} $\uparrow$ & \textbf{ mAP} $\uparrow$  \\ \hline
    \multirow{5}{*}{{ \rotatebox[origin=l]{90}{ \small BDD100K}}} & \multirow{5}{*}{{ \rotatebox[origin=l]{90}{\small nuImages}}}
        & GAN~\cite{lee2018training}&	83.65 & 70.39 & 31.5 \\
      &  & Energy~\cite{liu2020energy}
    &  81.62 & 69.43 & 32.0 \\ 
  & & CSI~\cite{tack2020csi}& 80.00 & 74.91 & 31.8 \\
        & & STUD~\cite{du2022stud}&	79.75 & 76.55 & 32.3\\
        & & FFS (ours)  & \textbf{76.68} & \textbf{77.53} & \textbf{36.2} \\
        \hline
    \multirow{5}{*}{{ \rotatebox[origin=l]{90}{\small Youtube-VIS} }} & \multirow{5}{*}{{ \rotatebox[origin=l]{90}{\small MS-COCO} }}
        & GAN~\cite{lee2018training}&	85.75 & 72.95 & 25.5\\
      & & Energy~\cite{liu2020energy}
    & 88.54 & 67.83 & 26.7 \\ 
  & & CSI~\cite{tack2020csi}&82.43 & 71.81 & 24.2 \\
& & STUD~\cite{du2022stud}&	81.14 & 74.82 & 27.2\\
& & FFS (ours) & \textbf{83.06} & \textbf{76.37} & \textbf{27.6} \\
        \bottomrule
        \end{tabular}}
        \vspace{-0.2cm}
        \caption[]{\small Main results on video datasets
with Faster R-CNN model and RegNetX-4.0GF~\cite{DBLP:conf/cvpr/RadosavovicKGHD20} as the backbone network. }
        \label{tab:sota_video}
        \vspace{-0.4em}
\end{table}

In Table~\ref{tab:sota_image},  we report a 7.58\% and a 6.73\% decrease in FPR95 for $\texttt{FFS}$ compared to VOS when evaluated on MS-COCO and OpenImages, respectively. Additionally, we note an increase in AUROC values when compared with VOS and assessed on MS-COCO and OpenImages. We also obtained higher ID detection performance and a faster training time of 2.18 hr compared to 2.43 hr for VOS on the same hardware and batch size. In Table~\ref{tab:sota_video}, $\texttt{FFS}$ outperforms STUD, where we report a 3.85\% decrease in FPR95 and a 1.28\% increase in AUROC compared to STUD when the ID is BDD100K, and the OD is nuImages. We also register a significant 12\% increase in mAP performance compared to STUD. Our training time is 2.7 hrs compared to 11 hrs by STUD for the BDD100K dataset on the same hardware. \texttt{FFS} also improves STUD results for the Youtube-VIS as ID and MS-COCO as OD while maintaining high ID detection performance. For this dataset, we report a lower training time of 8.2 hrs compared to 11.3 hrs by STUD.  

\begin{table}[!htb]
\centering
\scalebox{0.75}{\begin{tabular}{c|c|c|c|c|c}
    \toprule
    {\textbf{ID}}  & 
   {\textbf{Method}}  &\textbf{FPR95} $\downarrow$  & \textbf{AUROC} $\uparrow$ & \textbf{ mAP} $\uparrow$  & \textbf{Time (h)}\\ \hline
    \multirow{2}{*}{{{ \footnotesize PASCAL-VOC}}}
     & VOS~\cite{du2022vos}&	49.67 & 88.43 & 48.91 & 2.37\\
        & FFS (ours)  & \textbf{43.12} & \textbf{89.84} & \textbf{51.73} & \textbf{2.12}  \\
        \hline
    \multirow{2}{*}{{{\footnotesize Youtube-VIS}}} 
&  STUD~\cite{du2022stud}&	83.93 & 74.54 & 24.42 & 11.03\\
&  FFS (ours) & \textbf{81.90} & \textbf{76.35} & \textbf{25.10} & \textbf{8.23} \\
        \bottomrule
        \end{tabular}}
        \vspace{-0.2cm}
        \caption[]{\small Performance evaluation with ResNet-50 backbone.}
        \label{tab:resnet}
        \vspace{-0.4em}
\end{table}

In Table~\ref{tab:resnet}, we also show the performance while using ResNet$-50$~\cite{DBLP:conf/cvpr/HeZRS16} as an alternative backbone and compare our results with VOS~\cite{du2022vos} and STUD~\cite{du2022stud} with MS-COCO as the outlier dataset. The results show that using ResNet-50 as the backbone achieves state-of-the-art ID and OD detection performance with our framework. Overall, Tables~\ref{tab:sota_image},~\ref{tab:sota_video}~and~\ref{tab:resnet}  demonstrate the effectiveness of \texttt{FFS} for OD detection for images and videos with lower training time and high ID detection performance. 

 \begin{figure*}[!htb]
    \centering
    \includegraphics[width=1.0\linewidth]{ood_images-compressed.pdf}
    \vspace{-2em}
    \caption{\small Visualization of OD detection on MS-COCO and OpenImages with VOS~\cite{du2022vos} (first and third row) and \texttt{FFS} (ours) (second and fourth row) trained on PASCAL-VOC 2012  images. Ideally, an object in the images should be labeled as OOD with a green bounding box. }
    \label{fig:visual}
    \vspace{-1.2em}
\end{figure*}

\begin{figure}[!t]
    \centering
    \includegraphics[width=1.0\linewidth]{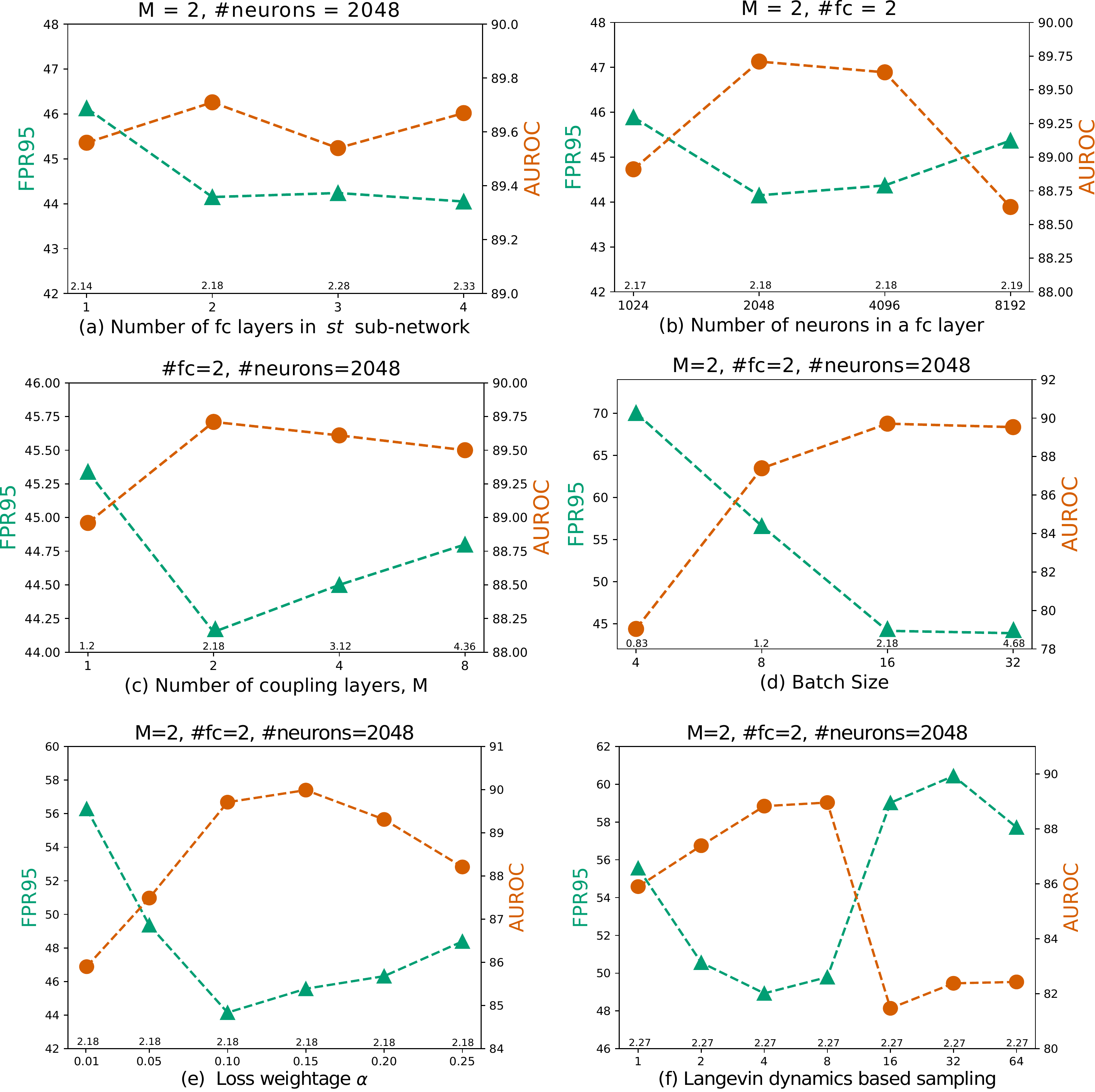}
    \caption{\small Validation  of the flow architecture, loss weightage $\alpha$,  batch size, and projection sampling. The training time (in hours) is shown above the horizontal x axis in the plots.}
    \label{fig:lineplots}
    \vspace{-0.4em}
\end{figure}

\subsection{Ablation Studies}
\label{ablation}
In this section, we perform an exhaustive study of meta-parameters that affects the performance of $\texttt{FFS}$. We also qualitatively compare OD detection results with the VOS approach. For uniformity, we fix the ID dataset as PASCAL-VOC and OD dataset as MS-COCO unless stated otherwise in the experiments.

\textbf{VOS plus rejection sampling:} Sampling outliers from a class conditional Gaussian (as performed in VOS~\cite{du2022vos}) does not ensure that such an outlier will have lower likelihoods for other class Gaussians. Therefore, we compute the log-likelihoods of the outliers using all class Gaussians and reject those with a higher likelihood from at least one other class compared to its generator class. Table~\ref{tab:rejection} shows the results for VOS~\cite{du2022vos}, VOS plus rejection sampling (named as VOS+), and \texttt{FFS}. VOS+ enhances the OD detection compared to VOS~\cite{du2022vos} due to selecting more effective outliers, but it is still worse than \texttt{FFS}. Furthermore, VOS+'s training time is much higher due to the iterative process of computing log-likelihoods through each Gaussian class. 

 \begin{table}[!ht]
\centering
\scalebox{0.78}{\begin{tabular}{lcccc}
    \toprule

   {\textbf{Method }}  &\textbf{FPR95} $\downarrow$  & \textbf{AUROC} $\uparrow$ & \textbf{ mAP (ID)}$\uparrow$ & \textbf{Time (h)} \\ \hline
        & \multicolumn{2}{c}{OD: MS-COCO / OpenImages} \\
         \cline{2-3}
  VOS~\cite{du2022vos}&	48.22 / 52.74 & 89.04 / 85.97 & 51.50 & 2.43\\
    VOS+&	45.59 / 47.96 & 89.40 / 87.19 & 51.74 & 7.38\\
        FFS (ours)  & \textbf{44.15} / \textbf{45.08} & \textbf{89.71} / \textbf{88.29}  & \textbf{51.80} & \textbf{2.18} \\
        \bottomrule
\end{tabular}}
\vspace{-0.2cm}
        \caption[]{\small Comparison of VOS, VOS+, and \texttt{FFS} (ours). The baseline mAP of a closed-set object detector is 51.35.}
        \label{tab:rejection}
        \vspace{-0.4em}
\end{table}

 \textbf{Normalizing Flow:} We conducted a thorough study on several types of normalizing flow models. We selected NICE~\cite{DBLP:journals/corr/DinhKB14}, RealNVP~\cite{dinh2017density}, Glow~\cite{kingma2018glow} and GIN~\cite{Sorrenson2020Disentanglement} in the chronological order for this experiment. We fixed the same coupling layers and each model's sub-network $s$ and $t$ configuration for an equivalent comparison. Table~\ref{tab:flow_arch} shows that Glow outperforms NICE, RealNVP and GIN in terms of both FPR95 and AUROC scores. GIN performs best for ID detection but has a worse OD detection performance. As the primary task of $\texttt{FFS}$ is OD detection, we chose Glow as our flow model for all our experiments. 

  \vspace{-0.5em}

\begin{table}[!htb]
\centering
\scalebox{0.85}{
\begin{tabular}{l|c|c|c|c}
    \toprule \multirow{1}{0.28\linewidth}{\textbf{Flow model}}  & \textbf{FPR95} $\downarrow$ & \textbf{AUROC} $\uparrow$ &  \textbf{mAP}  $\uparrow$ & \textbf{Time (h)} \\
    \hline
        NICE~\cite{DBLP:journals/corr/DinhKB14} &  48.23 & 88.46 & 51.81 & 2.23 \\
    RealNVP~\cite{dinh2017density} &  45.76 & 88.95 & 51.84 & 2.28 \\
    Glow~\cite{kingma2018glow} &   \textbf{44.15} & \textbf{89.71} & 51.80 & \textbf{2.18} \\    
    GIN~\cite{Sorrenson2020Disentanglement} &   48.31 & 88.36 & \textbf{52.00} & 2.25 \\
        \bottomrule
\end{tabular}}
        \vspace{-0.2cm}
        \caption[]{\small {Validation of the Normalizing Flow models. } }
        \label{tab:flow_arch}
        \vspace{-0.4em}
\end{table}

\vspace{-0.4em}
Figure~\ref{fig:lineplots} (c) shows the effect of the number of coupling layers $M$ on OD detection performance.~$M$ = $2$ produces the best results as the inlier features fit a suitable number of flow parameters.~However, the OD detection performance degrades for $M$ $>$ $2$, and the training time significantly increases due to more trainable parameters in the network.~Adding more trainable parameters requires more training data for effective training; otherwise, it might lead to overfitting and a more complex optimization, which should explain the degradation in performance. Hence, we fix $M = 2$, and in Figure~\ref{fig:lineplots} (a), we study the effect of the number of fully-connected (fc) layers within $s$ and $t$ sub-networks of each coupling layer. There is a slight improvement in OD detection performance when two fc layers are employed. In Figure~\ref{fig:lineplots} (b), we fix the number of coupling and fc layers in each $s$ and $t$ sub-networks as two and adjust the number of neurons in each fc layer. We report a degradation in OD detection performance above 2048 neurons.

 \small
 \begin{table}[!ht]
\centering
\scalebox{0.9}{\begin{tabular}{c|c|c|c|c}
    \toprule
   {\textbf{$\mathcal{L}_{reg}$}}  &\textbf{FPR95} $\downarrow$  & \textbf{AUROC} $\uparrow$ & \textbf{mAP} $\uparrow$ & \textbf{Time (h)} \\ \hline
    CE &	69.92  & 84.72  & 48.70 & 2.27 \\
     JSD  &	52.83  & 85.79 & 48.60 & 2.25\\
     Hinge  & 51.32  & 86.34 & 46.54 & 2.19  \\
     BCE (Ours)  & \textbf{44.15}  & \textbf{89.71}  & \textbf{51.80} & \textbf{2.18} \\
        \bottomrule
\end{tabular}}
        \vspace{-0.2cm}
        \caption[]{\small Validation of $\mathcal{L}_{reg}$ on OD detection performance.}
        \label{tab:lreg}
        \vspace{-0.4em}
\end{table}
\normalsize

\textbf{Regularization loss $\mathcal{L}_{reg}$ and its weightage $\alpha$:}
We studied several regularization losses for our \texttt{FFS} framework. Firstly, we used a simple multi-class cross-entropy loss (CE) on the classification logits without including the energy module. Secondly, we evaluated Jensen-Shannon divergence (JSD) loss, where we minimized the divergence between the distribution of the classification logits of the outlier samples and  uniform distribution. In Table~\ref{tab:lreg}, we show that our BCE based $\mathcal{L}_{reg}$ loss outperforms all other loss functions in terms of OD detection performance. The ID detection performance is also higher, even though the training time is comparable. In Figure~\ref{fig:lineplots} (e),  we demonstrate the effect of changing the weightage $\alpha$ of $\mathcal{L}_{reg}$ for PASCAL-VOC as ID and MS-COCO as OD. As $\alpha$ increases, the OD detection performance improves. However, an even higher weightage degrades the OD detection performance, due to which a careful selection of $\alpha$ is desirable.

\textbf{Projection Sampling:}
\label{langevin}
Figure~\ref{fig:lineplots} (f) shows the results after varying synthetic outliers. The OD detection improves as we increase the number of sampled outliers $o_s$ but degrades with much higher values of $s$. The coverage of log-likelihoods from many sampled outliers broadens even though the average log-likelihood is similar to the likelihood threshold $\delta$. Hence, some outliers may lie inside, while others exist far outside the decision boundary leading to bad regularization. We report that rejection sampling achieves better OD detection performance when compared to the best results of projection sampling. Since projection sampling enforces the outliers to be close to the decision boundary, it may be insufficient to estimate the entire outlier manifold. In contrast, the rejection sampling synthesizes outliers near the decision boundary and in the proper outlier manifold, leading to a better model regularization.

\textbf{Qualitative analysis:} Figure~\ref{fig:visual} shows the OD detection results on MS-COCO and OpenImages with \texttt{FFS} and VOS trained on PASCAL-VOC. The results show that \texttt{FFS} outperforms VOS in recognizing outlier objects. Additionally, our method's softmax confidence score is lower than VOS when both methods misclassify the outlier objects as inliers.


\vspace{-0.5em}
\section{Conclusion}
We presented a new methodology for outlier-aware object detection that learns a combined data distribution of all inlier object classes. By sampling from the low-likelihood region of a jointly trained normalizing flow model, our approach generates suitable synthetic outlier samples
for training the outlier detection head
of our compound model. In contrast, previous approaches model inliers with class-conditional Gaussians, resulting in outlier samples 
that may have a high likelihood
for some other class. We report state-of-the-art results for image and video datasets with standard outlier detection metrics while decreasing the model training time. We also show that simple rejection sampling
contributes more useful synthetic outliers
than projection sampling with 
a fixed likelihood threshold. In future work, one could validate this finding on other OD detectors. Our idea also extends to developing an outlier-aware instance segmentation model and an active learning scheme to correct the failure modes in weakly-supervised object detectors. 
\vspace{-1em}
\paragraph{Acknowledgement:} This work primarily received funding from the German Federal Ministry of Education and Research (BMBF) under Software  Campus (grant 01IS17044) and was supported by the Center for Scalable Data Analytics and Artificial Intelligence (ScaDS.AI) Dresden/Leipzig, Germany.  The  work  was also partially funded by DFG grant 389792660 as part of \href{https://perspicuous-computing.science}{TRR~248 -- CPEC}, Croatian Science Foundation (grant IP-2020-02-5851 ADEPT) and the Cluster of Excellence CeTI (EXC2050/1, grant 390696704). The authors gratefully acknowledge the Center for Information Services and HPC (ZIH) at TU Dresden for providing computing resources.

{\small
\bibliographystyle{ieee_fullname}
\bibliography{PaperForReview}
}

\newpage
\onecolumn
\appendix

\begin{center}
    \Large{\textbf{Supplementary Material}}
\end{center}

\section{Glossary}

\begin{tabular}{ m{3em} m{20cm} }
 $\mathcal{X}$ & The training dataset with $\mathcal{X}:= (x_1, x_2,..., x_N)$ containing $N$ images or videos  \\ 
 $\mathcal{Y}$  & The ground-truth labels with $\mathcal{Y}:= \{1, 2,..., K\}$ for $K+1$ inlier object classes \\  
$\mathcal{B}$ & The ground-truth coordinates of the bounding boxes with $\boldsymbol{b} \in \mathcal{B}$ \\
$l(x,\boldsymbol{\hat{b}})$  & The fixed-size box features of both inlier and background patches given the predicted bounding boxes $\boldsymbol{\hat{b}}$ \\
$l_{ID}(x,\boldsymbol{\hat{b}})$ & The fixed-size box features of only inlier  patches given the predicted bounding boxes $\boldsymbol{\hat{b}}$ \\
$\mathcal{L}_{det}$ & The standard object detection loss consisting of object classification and bounding-box regression losses \\
$\mathcal{L}_{nll}$ &  The negative log-likelihood loss to train the normalizing flow model on inlier features \\
$\mathcal{L}_{reg}$ & The regularization loss for discriminative training using inlier and synthetic outlier features  \\
$\xi$ & The energy-based threshold, when 95\% of inlier objects in the validation set are correctly detected \\
$h$ &  The classification head in the ROI head module of the Faster R-CNN architecture \\
$f$ &  The normalizing flow network to maximize the likelihood of inlier features \\
$\mathcal{Z}$ & The latent space of the flow network $f$ defined as a multivariate Gaussian with zero mean and unit variance \\
$\Phi$ &  The binary classifier to differentiate the inlier from the synthesized outlier features via discriminative training \\
$\theta$ &  The learnable parameters of the standard object detector, i.e. Faster R-CNN \\
$\gamma$ &  The learnable parameters of the normalizing flow $f$ \\
$\psi$ &  The learnable parameters of the binary classifier $\Phi$ \\
$g_k$ &   The generated synthetic features after randomly sampling $k$ samples from flow's latent space \\
$o_s$ &   The selected synthetic outlier features from the generated features $g_k$ such that $o_s \subset g_k$ \\
$\tau$ &   The step size of the gradient descent for the projection sampling based outlier synthesis \\
$\delta$ &   The log-likelihood  threshold to determine the synthesized outlier features based on projection sampling \\
$E(h(.))$ &   The energy-score calculated from the output of the classification head $h$ \\
$T$ & The temperature coefficient to compute the energy score $E(h(.))$ \\
$\alpha$ & The weightage of the regularization loss $\mathcal{L}_{reg}$ in the overall training objective \\
$\beta$ & The weightage of the negative log-likelihood loss $\mathcal{L}_{nll}$ in the overall training objective    
\end{tabular}

\section{Visualization of inlier and synthesized outlier features}
\begin{figure}[ht]
    \centering
    \begin{subfigure}[width=1.0\textwidth]{0.49\textwidth}
    \includegraphics[width=1.0\linewidth]{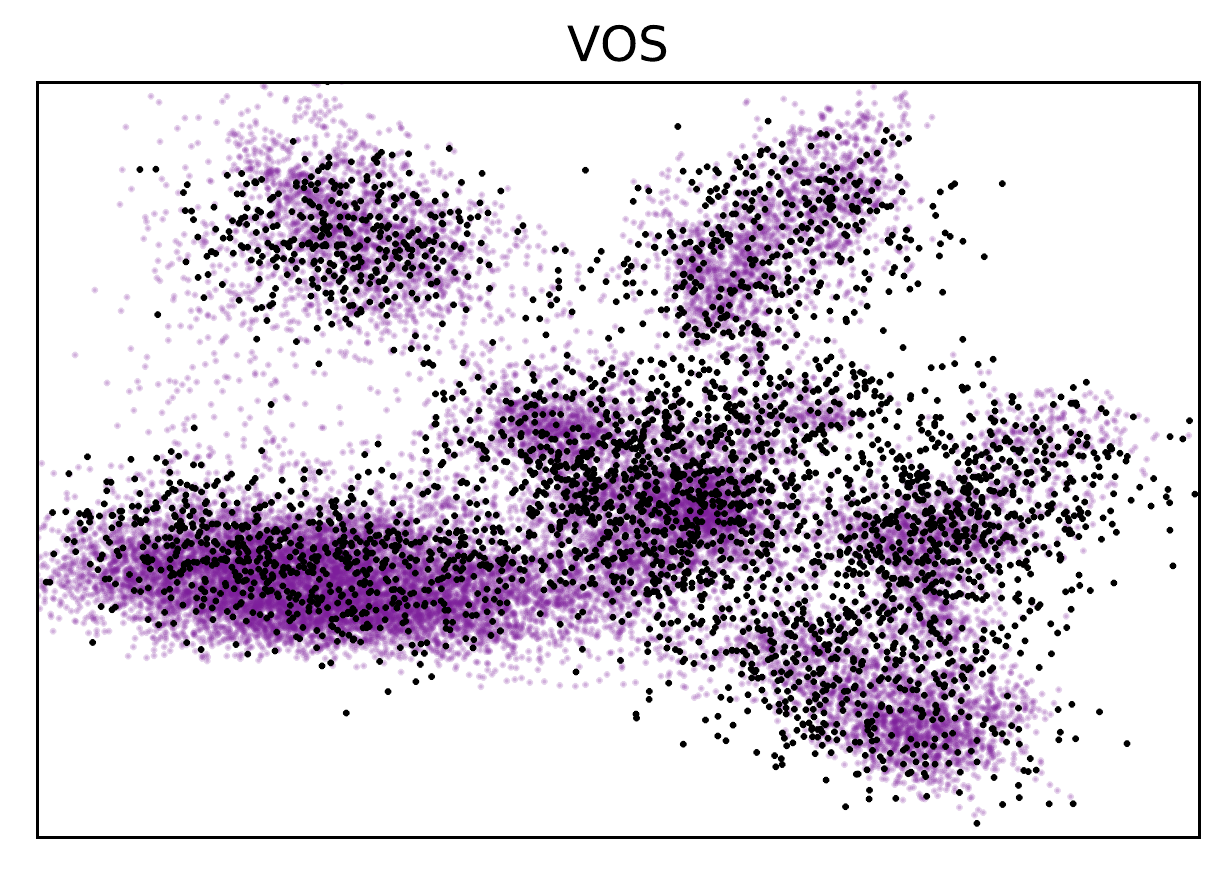}
     \end{subfigure}
    \begin{subfigure}[width=1.0\textwidth]{0.49\textwidth}
     \includegraphics[width=1.0\linewidth]{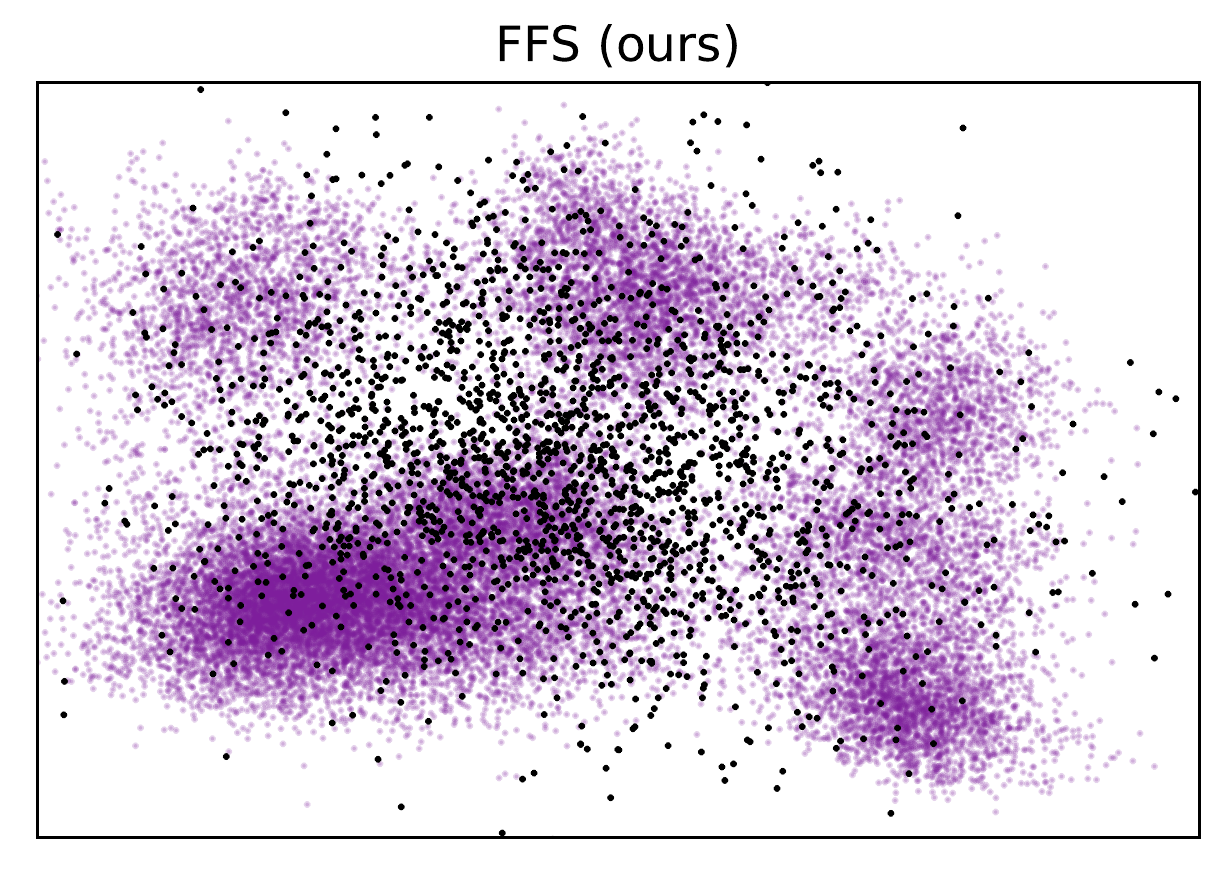}
     \end{subfigure}
    \caption{\small PCA visualization of synthesized outliers (in black) and inlier features (in color) of all 20 PASCAL-VOC object classes.}
    \label{fig:scatter}
    \vspace{-0.4em}
\end{figure}

We provide the visualization of inlier features (in color) of PASCAL-VOC along with the synthesized outliers (in black) after reducing the number of feature embeddings using Principal Component Analysis (PCA). We  compare the results from VOS~\cite{du2022vos} and our \texttt{FFS} approach in Figure~\ref{fig:scatter}. It is noticeable that VOS synthesizes outliers separately for each inlier class. In contrast, our approach synthesizes outliers after estimating the accurate data distribution of all inlier classes using the normalizing flow model, thereby leading to more effective regularization.

\section{Step size $\tau$ for projection sampling}
In Table~\ref{tab:projecton}, we show the OD detection results of \texttt{FFS} approach for varying step size $\tau$ when projection sampling was used to synthesize outliers precisely at the decision boundary. For this experiment, we fixed PASCAL-VOC as the inlier and MS-COCO as the outlier dataset, respectively. We analyzed the number $s$ of synthetic outliers, $o_s$, and $\tau$ to evaluate the performance. It can be seen from the results that, irrespective of the step size $\tau$, the OD detection performance improves as we increase the number $s$ of synthesized outliers $o_s$. However, the performance gets worse when the $s$ is increased further.

\begin{table}[!htb]
\centering
\scalebox{1}{\begin{tabular}{|c|c|c|c|}
    \toprule
    $\#$ samples, $s$  & \multicolumn{3}{c|}{Step size, $\tau$}  \\
    \hline
      & $\tau$ = 1  & $\tau$ = 0.5 & $\tau$ = 0.1 \\ \hline
       & \multicolumn{3}{c|}{FPR95 $\downarrow$ / AUROC $\uparrow$}  \\ 
    \cline{2-4}
1 & 55.56 / 85.90 & 51.13 / 86.73 & 52.34 / 86.21\\
2 & 50.56 / 87.39 & 49.48 / 87.95 & 49.94 / 86.92\\
4 & \textbf{48.93} / 88.83 & \textbf{47.23} / 89.44 & \textbf{47.49} / \textbf{88.72} \\
8 & 49.79 / \textbf{88.96} & 48.28 / \textbf{89.61}  &  48.95 / 88.84\\
16 & 59.02 / 81.47 & 59.60 / 82.09 & 59.32 / 83.54 \\
32 & 60.44 / 82.38 & 57.92 / 81.99 & 58.54 / 82.65 \\
64 &  57.72 / 82.43 & 58.67 / 82.22 & 59.91 / 83.43 \\
        \bottomrule
        \end{tabular}}
        \vspace{-0.2cm}
        \caption[]{\small OD detection results after varying the step size $\tau$ of our projection sampling based strategy to synthesize outliers.}
        \label{tab:projecton}
        \vspace{-0.4em}
\end{table}

\section{Datasets}
Our experimental setup consists of three inlier and outlier datasets, where we train \texttt{FFS} on both image and video-based inlier datasets. The image-based inlier dataset, i.e., PASCAL-VOC, does not contain the sequence of image frames in terms of time. In contrast, the video-based datasets, BDD100K and Youtube-VIS, are a sequence of image frames dependent on time.
We follow the experimental setup of VOS~\cite{du2022vos} and STUD~\cite{du2022stud} and utilize the datasets provided for training and inference of \texttt{FFS}. In addition, we show the object classes for each of the inlier datasets for a better interpretation of the subsequent visual results:

\begin{itemize}
    \item \textbf{\emph{PASCAL-VOC:}} There are 16,551 training and 4,952 validation images in the dataset. The object classes are \emph{person, bird, cat, cow, dog, horse, sheep, airplane, bicycle, boat, bus, car, motorcycle, train, bottle, chair, dining table, potted plant, couch and tv.}
    \item \textbf{\emph{BDD100K:}} There are 273,406 training and 39,973 validation images in the dataset. The object classes are  \emph{pedestrian, rider, car, truck, bus, train, motorcycle and bicycle.}
    \item \textbf{\emph{Youtube-VIS:}} There are 67,861 training and 21,889 validation images in the dataset. The object classes are \emph{airplane, bear, bird, boat, car, cat, cow, deer, dog, duck, earless seal, elephant, fish, flying disc, fox, frog, giant panda, giraffe, horse, leopard, lizard, monkey, motorbike, mouse, parrot, person, rabbit, shark, skateboard, snake, snowboard, squirrel, surfboard, tennis racket, tiger, train, truck, turtle, whale and zebra.}
\end{itemize}

We evaluate our trained \texttt{FFS} framework on three different outlier datasets, namely MS-COCO, OpenImages, and nuImages. There are 930 images in MS-COCO and 1,761 images in OpenImages datasets, so objects in these images do not fall into any of the inlier classes of PASCAL-VOC. Similarly, there are 2,100 images of the nuImages dataset for evaluating \texttt{FFS} trained on BDD100K and 28,922 images of MS-COCO for evaluating \texttt{FFS} trained on Youtube-VIS. In all outlier datasets, the objects present in the inlier dataset are mutually independent of objects in the outlier dataset.

\section{Visualization of OD detection results for video datasets}

\begin{figure}[!htb]
    \centering
    \includegraphics[width=0.95\textwidth]{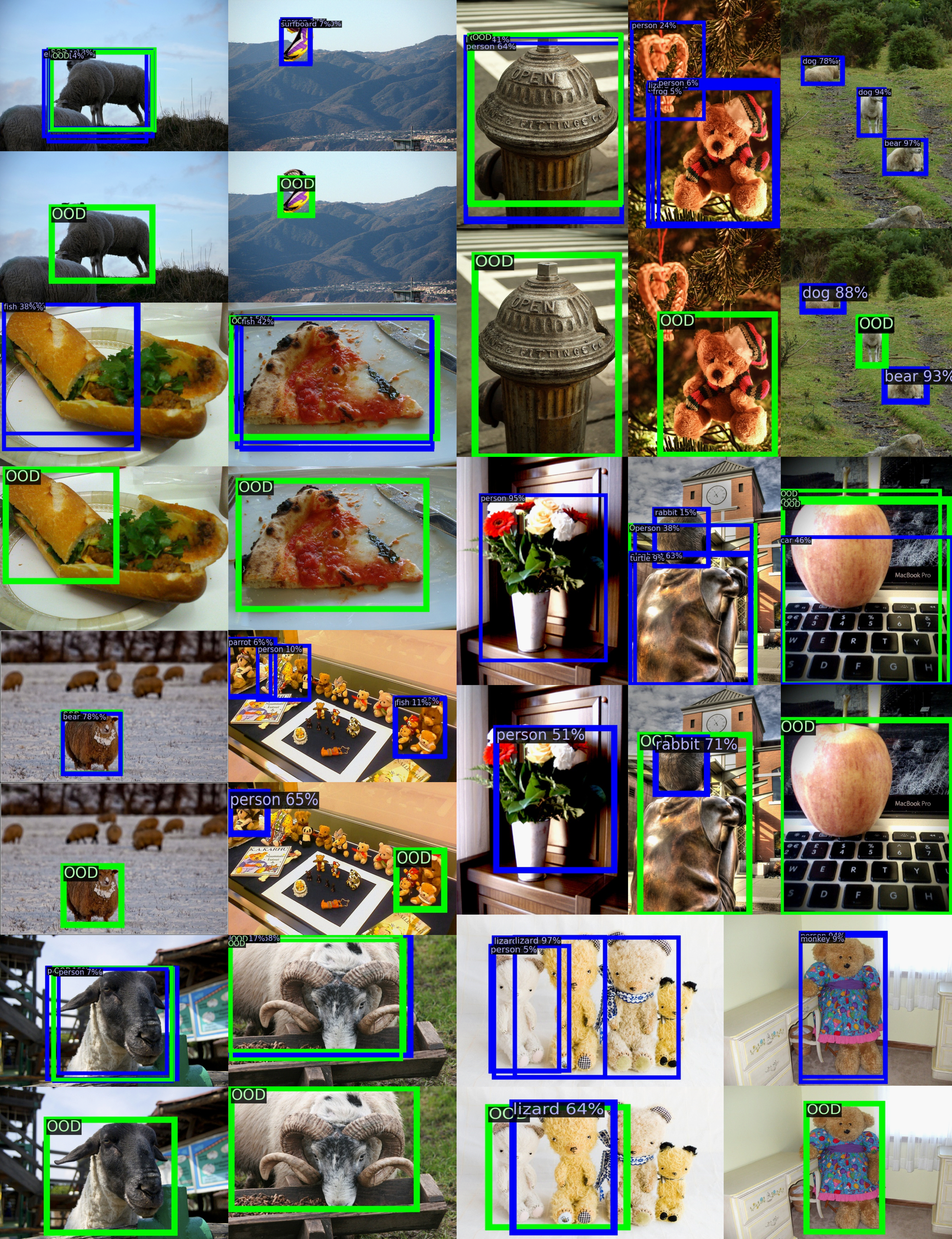}
    \caption{\small OD detection results on MS-COCO images when \texttt{FFS} was trained on Youtube-VIS dataset. In each image pair, the top image is the results from STUD~\cite{du2022stud}, and the bottom image is the results from \texttt{FFS}.}
\vspace{-2em}
\label{fig:supp1}
\end{figure}

\begin{figure}[!htb]
    \centering
    \includegraphics[width=1.0\textwidth]{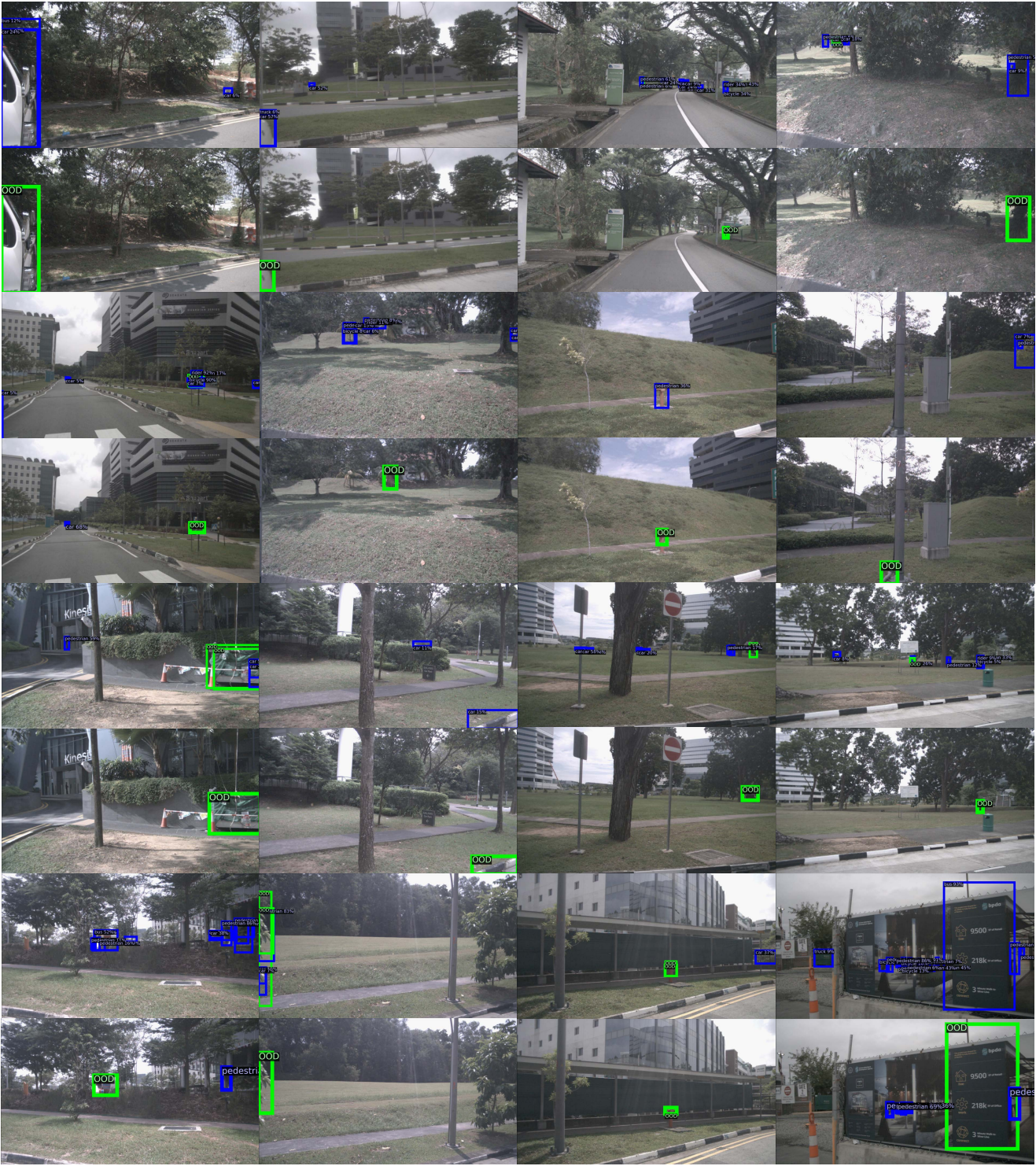}
    \caption{\small OD detection results on nuImages when \texttt{FFS} was trained on BDD100K dataset. In each image pair, the top image is the results from STUD~\cite{du2022stud}, and the bottom image is the results from \texttt{FFS}.}
\vspace{-2em}
\label{fig:supp2}
\end{figure}

In Figure~\ref{fig:supp1} and Figure~\ref{fig:supp2}, we show the OD detection results when \texttt{FFS}
was trained on video datasets. The results showcase that we obtain significantly better results in detecting outliers and reducing the number of incorrect bounding box predictions compared to STUD~\cite{du2022stud}. Additionally, for some image pairs, we reduce the model's confidence when the object was wrongly detected as an inlier by our approach and STUD~\cite{du2022stud}.

\vspace{-4em}

\end{document}